# Handwritten Arabic Character Recognition for Children Writing Using Convolutional Neural Network and Stroke Identification

**Mais Alheraki [1], Rawan Al-Matham[2] and Hend Al-Khalifa [3]**

[1]   iWAN Reseach Group, King Saud Univeristy; mais.alheraki@gmail.com
[2]   iWAN Reseach Group, King Saud Univeristy; r.almatham@gmail.com
[3]   iWAN Reseach Group, King Saud Univeristy; hendk@ksu.edu.sa

**Abstract:** Automatic Arabic handwritten recognition is one of the recently studied problems in the field of Machine Learning. Unlike Latin languages, Arabic is a Semitic language that forms a harder challenge, especially with variability of patterns caused by factors such as writer's age. Most of the studies focused on adults, with only one recent study on children. Moreover, much of the recent Machine Learning methods focused on using Convolutional Neural Networks, a powerful class of neural networks that can extract complex features from images. In this paper we propose a convolutional neural network (CNN) model that recognizes children handwriting with an accuracy of 91% on the Hijja dataset, a recent dataset built by collecting images of the Arabic characters written by children, and 97% on Arabic Handwritten Character Dataset. The results showed a good improvement over the proposed model from the Hijja dataset authors, yet it reveals a bigger challenge to solve for children's Arabic handwritten character recognition. Moreover, we proposed a new approach using multi models instead of single model based on the number of strokes in a character, and merged Hijja with AHCD which reached an averaged prediction accuracy of 96%.

## 1. Introduction

Handwriting recognition is one of computer vision problems. It is a process of automating the identification of handwritten script by a computer which transforms the text from a source such as documents or touch screens to a form that is understandable by the machine . The image can be offline input from a piece of paper or a photograph, and online if the source was digital such as touch screens [1].

The handwritten text for each language has many different patterns and styles from writer to writer. Many factors such as age, background, the native language, and mental state affect the patterns in any piece of handwritten text [2]. Automatic handwritten recognition is well investigated in the literature by using many methods revolving from machine learning such as: K-nearest Neighbors (KNNs) [3]Support Vector Machines (SVMs) and transfer learning  [4] [5]. Also, some studies used deep learning techniques such as Neural Networks (NNs) [6]. Recently, most of the studies have used Convolutional Neural Networks (CNNs) [5] [7][8] [9].

Latin languages have been intensively studied in the literature and achieved state-of-the-art results [10][3][11]. Nevertheless, the Arabic language still needs more investigation. Arabic is a Semitic language, it is the fourth most spoken language in the world [12]. Arabic has its own features that make it different from other languages including spelling, grammar, and pronunciation. Arabic writing is semi-cursive, and it is written from right to left. The Arabic alphabet contains 28 characters. Every character has many shapes depending on its position in the word. These aspects make the automatic handwritten recognition of the Arabic script harder than other languages. Many recent studies were conducted targeting Arabic handwritten recognition [13], [14], [15]. However, all of them focused on recognizing Arabic adult script except for [7]. They created Hijja dataset which was collected from 591 children from Arabic schools. Additionally, they proposed a CNN model to evaluate their dataset. Their achieved prediction accuracy was %87.

In this research, we aim to improve the prediction accuracy over Hijja dataset using CNN to have a more robust model that can recognize children's Arabic script. Our experiments will answer the following research questions:

(1) Using our newly proposed CNN architecture, can we enhance the accuracy for children's Arabic handwritten character recognition?

(2) Using character strokes as a filter, can we enhance the accuracy for children's Arabic handwritten character recognition?

The rest of the paper is organized as follows: in Section 2, we give a background about Arabic language and Arabic script, Optical character recognition, and Convolutional Neural Network. Section 3 provides an overview of the related work, our methodology including the proposed solution, the used dataset, and the experimental setup. Results from our experimentation are presented in Section 4. Section 5 discusses and analyzes the results; Section 6 concludes the paper with some future work.

## 2. Background

In this section, we present the necessary background information needed to explain the underlying concepts of this research including Arabic language and Arabic script, Optical character recognition, and Convolutional Neural Network.

*Arabic Language and Arabic Script*

Arabic is a Semitic language and the language of the Holy Qur'an. Almost 500 million people around the globe speak Arabic and it is the language officially used in many Arab countries with different dialects, and the formal written Arabic is Modern Standard Arabic (MSA). MSA is one form of classical Arabic, which is the language that was used in Qur'an, and it currently has a larger and modernized vocabulary. Because it is understood by almost everyone in the Arab world, it is being used as the formal language in media and education. Arabic has its own features that make it different from other languages including spelling, grammar, and pronunciation [2].

The calligraphic nature of Arabic script is different from other languages in many ways. Arabic writing is semi-cursive, and it is written from right to left. The Arabic alphabet is 28 characters. Their shapes change depending on the position in the word. There are 16 characters, which contain dots (one, two, or three) among the 28 of the Arabic alphabets. These dots appear either above or below the character. Some characters may have the same body but with different dots number and/or position, as shown in **Figure 1**.

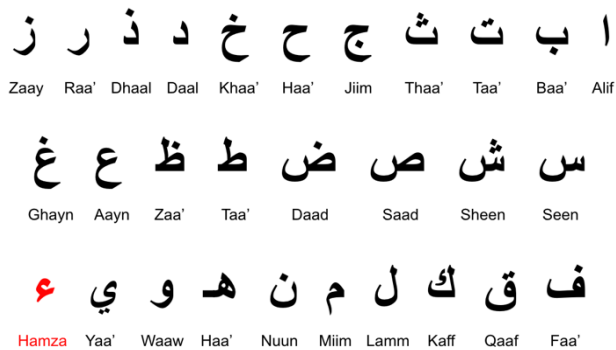

***Figure 1:*** *The Arabic characters, Hamza is colored by red as it is not part of the 28 alphabets*

Arabic characters have different shapes depending on their position in a word: initial, medial, final, or stand-alone. The initial and medial shapes are typically similar and so are the final and stand-alone shapes, see **Table 1**.

These Aspects make recognizing Arabic script tasks more challenging compared to Latin script. Because of that, there are fewer resources created for this task and thus the-state-of-art is less advanced. Nevertheless, there were some efforts to recognize Arabic handwriting in the last few years which is covered in section 3.

*Optical character recognition*

Optical Character Recognition (OCR) is a pattern recognition problem that takes printed or handwritten text as an input and creates an editable machine-understandable format of text extracted from the scanned image. OCR can be used in many applications such as advanced document scanning, business applications, electronic data searching, data entry, systems for visually challenged persons, document verification, document automation, data mining, biometrics, text storage optimization, etc. [16].

OCR is divided into offline and online recognition. It depends on the type of data that is used as an input for recognition. In offline recognition, the input is only an image of the handwriting text with less information. In Online recognition, a special input device, e.g., an electronic pen, tracks the movement of the pen during the writing process. Offline recognition is usually more difficult and challenging than online recognition [1].

*Table 1: Isolated, initial, medial, and final shapes of the Arabic characters*

| | b | n | y | J | Y | m | f | q | s | S | T | h | k | l | A | d | r | w | ʻ |
|---|---|---|---|---|---|---|---|---|---|---|---|---|---|---|---|---|---|---|---|
| Isolated | ب | ن | ي | ج | غ | م | ف | ق | س | ص | ط | ه | ك | ل | ا | د | ر | و | ع |
| Initial | بـ | نـ | يـ | جـ | غـ | مـ | فـ | قـ | سـ | صـ | طـ | هـ | كـ | لـ | | | | | |
| Medial | ـبـ | ـنـ | ـيـ | ـجـ | ـغـ | ـمـ | ـفـ | ـقـ | ـسـ | ـصـ | ـطـ | ـهـ | ـكـ | ـلـ | ـا | ـد | ـر | ـو | ء |
| Final | ـب | ـن | ـي | ـج | ـغ | ـم | ـف | ـق | ـس | ـص | ـط | ـه | ـك | ـل | | | | | |

*Convolutional Neural Network*

Convolutional Neural Network is a special type of neural networks that is widely used in deep learning to extract features from visual data. CNNs proved the state-of-art results in many image classification problems. At any computer vision problem, CNNs are the best candidate for reaching considerably high accuracies compared to other machine learning (ML) algorithms. A CNN takes an image as an input, in the form of a 3D matrix, with width, height and channels, then applies several filters on this matrix; those filters are called kernels and have different types to extract different features on each convolutional layer. There are several layers in a CNN besides convolutions, which can be different from network to network, but the deeper a CNN the numerous weights it will have, so the pooling layer comes to reduce the size of convoluted layers, by either finding the max pixel value in a window or average of all values, this is important to decrease the computational resources needed in CNNs. Other layers include the activation layer, which could be ReLU or any other activation, and normalization [4].

## 3. Literature Review

In this section, we present some of the existing datasets that were used for Arabic handwritten recognition in the literature. Additionally, some of the carried-out literature used CNNs for Arabic handwriting recognition.

*Arabic Handwritten Recognition Datasets*

There are many datasets created for Arabic handwritten recognition. One of them is introduced in [17]. It contains 5,600 images written by 50 adult writers which includes a variety of shapes for each character. Then

DBAHCL dataset was introduced in [18]. It includes 9900 ligatures and 5500 characters written by 50 writers for handwritten characters and ligatures.

Another work was conducted to collect Arabic handwritten diacritics (DBAHD) [19]. Another dataset called AHCD was introduced in [20]. It includes 16800 characters were written by 60 adult writers. The character images only in isolated form [Arabic Handwritten Characters Recognition using Convolutional Neural Network] and it was used in many such as studies [20] [21] [7].

Lastly, Hijja dataset was introduced and experimented with in [7] which includes 47,434 characters written by 591 children. The characters are written in isolated and connected forms and it is the biggest existing dataset. We chose the last two dataset (Hijja and AHCD) in this research.

*CNN in Arabic Handwritten Recognition*

In this section we present some of the carried-out literature that used CNNs for Arabic handwriting recognition. In [22], Elleuch et al. investigated two types of neural networks for Arabic handwritten recognition. They are Deep Belief Network (DBN) and Convolutional Neural Networks (CNN). The two networks used a greedy layer-wise unsupervised learning algorithm for processing. The experiments were done on HACDB Dataset and CNN obtained the best results with 14.71 % error classification rate on the test set.

Similarly, In [9], El-Sawy et al. designed and optimized CNN classifier by working on the learning rate and activation function (ReLU function). Their experiments were done on AHCD dataset, and they achieved an accuracy of 94.9% on testing data.

Additionally, in [20], El-Sawy et al. aimed to recognize Arabic digits by using a CNN based on LeNet-5 to recognize Arabic digits. Their experiments were conducted on a MADBase database (Arabic handwritten digits images). Their model achieved high accuracy with 1% training misclassification error rate and 12 % testing miss classification error rate.

In [23], Amrouch et al. used CNN as an automatic feature extractor in the preprocessing stage and Hidden Markov Models (HMM) as recognizer. This made the feature extraction process easier, faster, and more accurate than the manual feature extraction. They achieved 89.23% accuracy.

In [15], Ashiquzzaman et al. developed a method for Handwritten Arabic Numerals using CNN classifier. They achieved 99.4% of accuracy with dropout and data augmentation. Their experiments were done on CMATERDB dataset, and they improved the accuracy by inverting the images colors such that the number is white on a black background, this is observed from previous studies that it makes it easier to detect edges.

In [5], Soumia et. al. compared two approaches to Arabic handwritten character recognition. The first one using Conventional machine learning using the SVM classifier. The second one used Transfer learning with ResNet, Inception V3, and VGG16 models. They also proposed a new CNN architecture and tested it. The best accuracy results achieved with their CNN model, and it achieved 94.7%, 98.3%, and 95.2% on three databases OIHACDB-28, OIHACDB-40, and AIA9K respectively. However, all of the previous studies were focused on recognizing adult's Arabic handwriting script.

In [24], Ahmed et. al. presented CNN context based accitectre to recognise Arabic letters, words, and digits. They expermented with MADBase, CMATERDB, HACDB and SUST-ALT datasets. They aimed to reach the higset possple testing accuracy and they acived 99% for digits, 99% for letters and 99% for words on 99% daaset. However, their preposed model was desighned for offline recognetion while we aim to devlope a ruobost a lightwhigt model tha can be used for online recognetion for Arabic childern handwritten writing in reallife senarios.

In [25], Balaha et al. preposed two different approaches. The first one using 14 different CNN architectures on HMBD dataset and the best-aquired testing accuracy is 91.96%. Their sconed approach was using transfer learning (TF) and genetic algorithm (GA) approach named "HMB-AHCR-DLGA" was preposed to optimize the training parameters and hyperparameters in the recognition phase. The pre-trained CNN models (VGG16, VGG19, and MobileNetV2) are used in the second approach. The highest aquired testing accuracy was 92.88%.

There was one study focused on children's Arabic handwriting scrip [7], Altwaijry et al. described a new collected dataset of Arabic letters written by kids aged 7–12. The dataset is called Hijja, and it includes 47,434 characters written by 591 participants. Also, they proposed a CNN model for Arabic handwritten recognition which was trained and tested on Hijja and Arabic Handwritten Character Dataset (AHCD) dataset. Their model achieved accuracies of 97% and 88% on the AHCD dataset and the Hijja dataset, respectively.

All these studies proved that CNNs are the most suitable approach for Arabic handwritten recognition due to its power in automatic feature extraction which make it able to understand the difficult patterns in handwritten text. Therefore, we decided to use it in this research with a new proposed architecture to enhance the prediction accuracy on Hijja dataset to recognize children's handwriting script. We also decided to invert the images before feeding them to the model since it had positive improvements in [15].

## 4. Methodology

To build a model that can recognize handwritten characters, we decided to go with a deep learning approach using Convolutional Neural Networks (CNN). CNN has proved to be strong in automatic feature extraction and has reached state of the art in many image classification problems.

In this section, we present a CNN archeticture and two training approaches, with the goal of further enriching the results carried out in the literature. In the first approach, we go with a single model trained on multiple datasets. The second approach is based on the number of strokes in each character, where we divided the characters into 4 groups, as shown in **Table 2**. Each group was used to train a different model, then the number of strokes will be used as a filtration step before choosing which model will make the prediction. Therefore, we refer to them as the single-model approach, and multi-model approach respectively.

**Table 2** shows what characters belong to each group, it's also seen that some characters belong to 2 groups, columns represent groups, rows represent the handwriting style which could cause the number of strokes to be different. A stroke in our work starts from the writer's stylus, pen, or whatever method used for typing, touching the surface to form one part of the character, that part could be the main body of the character, or a dot, until the writer lefts up their hand off the surface.

*Table 2: One group = one trained model only on the characters in the group.*

|  | Group 1 | Group 2 | Group 3 | Group 4 |
|---|---|---|---|---|
| **Basic Naskh** نسخ | ا، ح، د، ر، س، ص، ط، ع، ل، م، هـ، و، ء | أ، ب، ج، خ، ذ، ز، ض، ظ، غ، ف، كـ، ن | ت، ق، ي | ث، ش |
| **Other Handwriting Styles** |  | ت، ث، ش، ط، ق، ي | ظ |  |

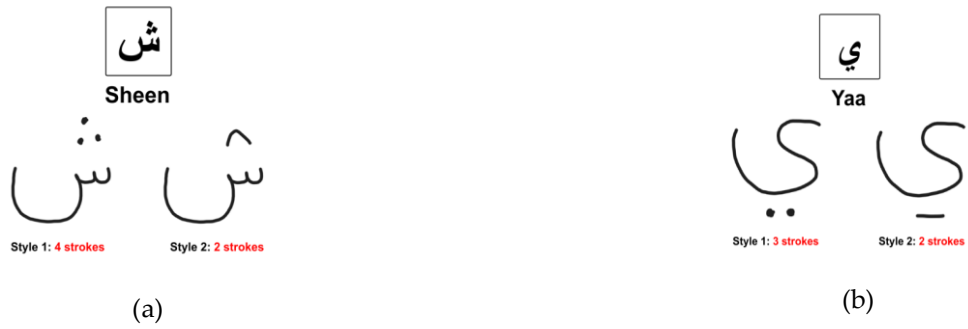

(a)                                                    (b)

*Figure 2: (a) Sheen belongs to Group 4 and 2, the dots could be written separately which make it 3 strokes or merged into a curve which make it one stroke. (b) Yaa belongs to Group 3 and 2, same case caused by dots, either separated or merged into a dash.*

### 3.1. Datasets

There are a couple of datasets for Arabic handwritten characters, as discussed in previous sections, in this experiment we used Hijja, AHCD, and a third dataset built by merging Hijja with AHCD into one new dataset, we call it Hijja-AHCD.

Hijja dataset [7] consists of the 28 letters in the Arabic alphabet, in addition to the form of isolated Hamza (ء), which makes it 29 total classes. The data was collected from 591 children in Arabic-speaking schools located in Riyadh, Saudi Arabia. It contains a total of 47,434 characters. This dataset includes both the isolated and connected forms of Arabic characters, as in the Arabic script, characters could have up to 3 different forms based on their location in the word. This variation made it perfect for this experiment as it will allow the network to recognize Arabic handwritten characters in all their forms.

Arabic Handwritten Characters Dataset (AHCD) [20] is a publicly available dataset, which contains 16,800 characters written by 60 participants with ages ranging between 19 and 40. Unlike Hijja, the age range makes the style of the handwritten characters different, it's more clear and easier to recognize by the human eye. The total number of classes is 28, matching the number of characters in the Arabic alphabet. AHCD only contains the isolated forms of each character.

The third dataset was constructed by augmenting Hijja with AHCD, both has grey images with 32 by 32 resolution, which made the merging easy and doable. We did that by using pandas, as both datasets were available in CSV format, we merged them into a new dataset that also has the CSV format and called it Hijja-AHCD. The concatenation wasn't done programmatically before each experiment, instead we chose to form a new dataset so that carrying out different experiments is easier.

### 3.1.1. Data Preprocessing

When we initially loaded AHCD, and visualized a subset of it, the images were incorrect as it was rotated 90 degrees to the left and flipped vertically, as shown in **Figure 2:** (a) Sheen belongs to Group 4 and 2, the dots could be written separately which make it 3 strokes or merged into a curve which make it one stroke. (b) Yaa belongs to Group 3 and 2, same case caused by dots, either separated or merged into a dash. **Figure 3: (a)** is showing a subset from AHCD without any modification, **(b)** is the same image after transpose. We used NumPy to transpose each image, and then save the data back to CSV, so that we don't bother later in repeating this step for each experiment. Note that the step of constructing Hijja-AHCD was done with the transposed version of AHCD.

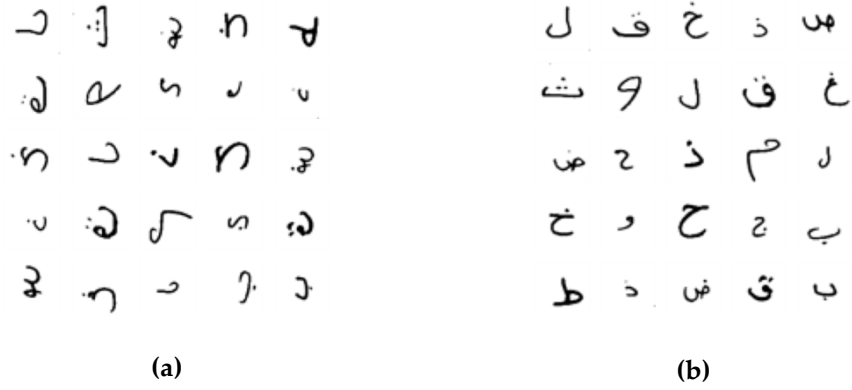

**(a)**                                                                 **(b)**

*Figure 3:* (a) is showing a subset from AHCD without any modification, (b) is the same image after transpose

The last step before training was to invert the colors for all images, such that it has a black background and a white foreground. This step was inspired from [20] study where they had better results in Arabic digits recognition after they applied this technique.

## 3.2. Model Architecture

In this section we propose a CNN architecture, which uses many convolution layers, and ends with a classification feedforward network, **Figure 4** shows the full architecture.

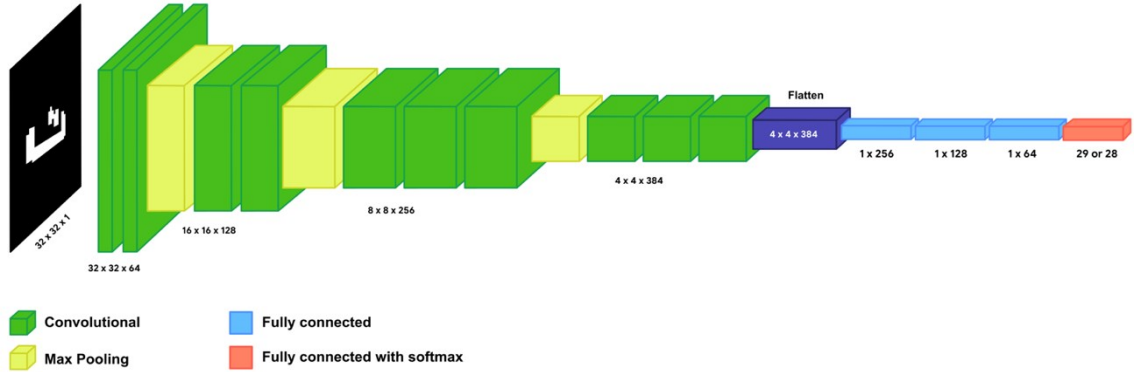

*Figure 4:* The architecture of the proposed CNN model

We start with an input image of 32 by 32 size and 1 grey channel. The model consists of 4 convolutional blocks, the first two blocks have 2 convolution layers with 64 filters and 128 respectively, followed by activation, then max pooling with size of 2, and finally a batch normalization step. The last two blocks have 3 convolution layers, with 256 and 384 filters respectively, followed by activation, max pooling, and batch normalization. Then, a flattening step is applied to prepare for feeding into the classifier. Instead of using one fully connected layer for classification, we use a feedforward network, consisting of 3 layers, with regularization using a dropout with 0.3 probability to avoid overfitting. HeUniform weight initializer was used instead of the default random initialization to initialize all weights of convolution and fully connected (FC) layers.

The feedforward network starts with a flattening step to turn the output from the feature extractor part into a vector with 1 dimension, that is, 4 by 4 by 384 long, and then feed it into the first FC layer, which will output 256 neurons, then 128, and lastly 64. The very last FC layer is a classifier with a softmax function that produces the class rates for the given labels, 28 in case of AHCD, and 29 in case of Hijja and Hijja-AHCD.

The activation function used in all the layers is LeayReLU, with a slope of 0.3. LeakyRelu is a variation of ReLU that was proposed to overcome the problem of vanishing gradient that arises in very deep networks when that uses ReLU. LeakyReLU keeps some of the negative data instead of assigning Zero to all of them, hence it's called leaky. Our experiments, as will be seen later, showed that LeakyReLU made a slight improvement on the results compared to the baseline.

To optimize the network, we used Categorical Cross Entropy for the loss function, since we have multiple class rates, and the labels were provided as one-hot encoded. Lastly, Adam optimizer was used to update the weights with an initial learning rate 0.001.

### 3.3 Experimental Setup

This section explains the carried-out experiments to recognize Arabic handwritten characters and goes over the training setup. We used Python language and Keras framework, with multiple programming libraries such as pandas, NumPy and scikit. All the work was done on **Google Colab with enabled GPU**. First, we followed several approaches for preprocessing. Then, we built the base CNN classification model as explained in section 3.2, which is used in all experiments, with fixed hyperparameters that we found the best after many experiments. Therefore, evaluating results would be based solely on the difference between each preprocessing and training approach.

### 3.3.1 Training Setup

To start the training, we need to have a validation set to help in detecting when the model starts overfitting on the training set. Cross validation was used as a technique to get the best validation set instead of splitting it manually with a fixed rate like in [4]. We used StratifiedKFold function from scikit-learn library to split the training set into training and validation sets, we set the number of splits to 5 and shuffle to true, meaning that the training loop will run 5 times each time with a different training and validation sets, so that we can pick the model with highest validation accuracy. Batch size is set to 128, and we trained for 30 epochs in all the five experiments, none of the models showed any improvement after epoch number 30.

During training, we used a learning rate scheduler, that is, Adam starts with an initial learning rate of 0.001, then during training, it optimizes the learning rate by itself, but we found, by experiment, that when adding a callback scheduler that is applied on the learning rate after Adam optimization on each epoch, the overall model accuracy improved. We set up this callback as a function that takes a learning rate as an input, then multiplies it by the exponential of -0.01.

### 3.3.2 Experiments

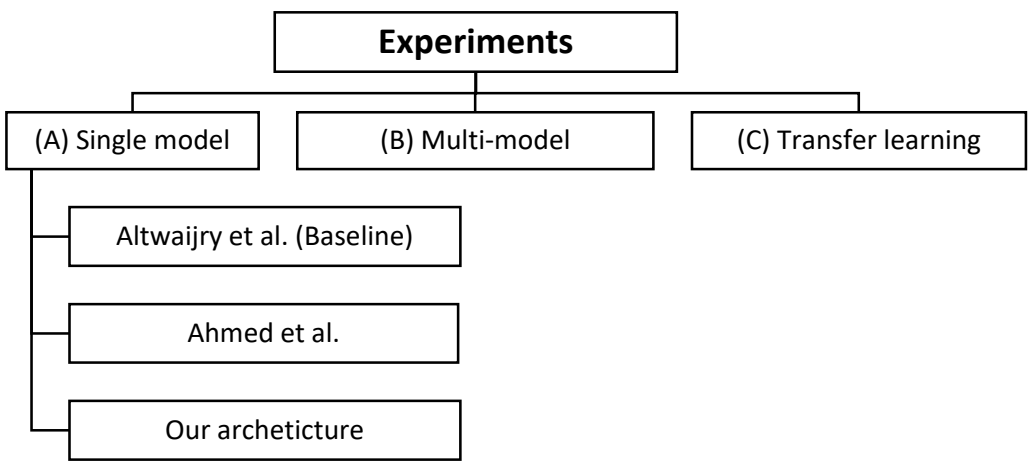

*Figure 5: experiments outline*

**Figure 5** shows all the carried experiments in this study. In (A), we conducted 3 experiments that falls under the single model approach, starting with the proposed model by Hijja dataset authors Altwaijry et al [7]., then the model proposed in the most recent study on Arabic handwritten recognition by Ahmed et al. [24], and finally our proposed model.

In (B), the multi-model approach, we used our proposed model to train and test on the Hijja-AHCD dataset. We started by filtering the characters based on number of strokes and splitted the data correspondingly. After we grouped the characters into 4 different groups based on their strokes; each group was used as a standalone dataset and is used to train. model, so we say this model is for group X. We refer to this approach as multi-model.

Lastly, we experimented transfer learning using EfficientNetV0 pre-trained model on Hijja, AHCD and Hijja-AHCD. Transfer learning helps in many problems by transferring the knowledge learned from one related or unrelated domain to another one which in many cases gives a higher convergence rate and results. There are many models trained on ImageNet dataset, which is a large dataset with 1000 classes of natural images, such as MobileNet, VGG19, EfficientNet and its variants, and many others. We picked EfficientNetV0 as it's one popular choice recently for vision problems due to its efficiency and relatively small size compared to others. The experiment started with substituing the top layer with our input size (32, 32, 3), but noticed here that EfficientNetV0 expects a 3 channels image so we needed to apply a preprocessing step to convert from Grayscale images to RGB. We also added a final classification layer to output 29 classes matching our data. The training will start with the ImageNet weights, and will train all EfficientNet layers. The rest of the setup is the same as our CNN model.

*3.4 Baseline and Evaluation Measures*

To evaluate how well our model performed in each experiment, we calculated the prediction accuracy, recall, precision, and F1 measures on the test set for each corresponding dataset. For (C) to be evaluated, we calculated the average prediction accuracy over the four models for each stroke-based character group. Same for precision, recall, and F1 score measures. We refer to Altwaijry et al. sa the bsaeline since their model was the first to be trained and tested on Hijja dataset.

3.4.1 Precision, Recall and F1 Score

Precision is the proportion of correctly classified characters from all characters in class X. While recall is the proportion of correctly classified characters from all characters in class X. F1 score is a function of precision and recall.

$$Precision = \frac{True\ Positive}{Trus\ Positive + False\ Positive}$$

$$Recall = \frac{True\ Positive}{Trus\ Positive + False\ Negative}$$

$$F1 = 2 \times \frac{Precision * Recall}{Precision + Recall}$$

**4 Results and Discussions**

In this section, we present and discuss the results of our experiments. **Table 3** shows the prediction accuracy for all of them. Refer to Appendix A for the full classification reports.

**Table 3:** *prediction accuracy for all experiments*

| | (A) Single CNN model | | | Transfer learning with EfficientNet V0 | Multi-model |
|---|---|---|---|---|---|
| | Altwaijry, et al. (Baseline) [7] | Ahmed, et al. [24] | Our model | | |
| Hijja Dataset | 87% | 90% | **91%** | 86% | - |
| AHCD Dataset | 96% | 94% | **97%** | 97% | - |
| Hijja-AHCD | - | 92% | 93% | 90% | **96%** |

Starting from first row, where the results for carrying out full experiments on Hijja dataset can be seen. Our model outperformed Altwaijry, et al. [7], Ahmed, et al.[24], and transfer learning with EfficientNetV0. In the second row, we used AHCD dataset. Our model obtained better results than both [7] Altwaijry, [24] Ahmed, but has a similar results with EfficientNetV0. Something that worth noting here is that there's a big difference in the accuracy of all single model experiments on Hijja versus AHCD datasets. We assume that this large difference in prediction accuracies come from the challenges Hijja dataset introduces. First, it contains different forms of each character, both isolated and connected, introducing more features and higher level of similarities between characters. Second, the characters were written by children. On the other hand, AHCD is written by adults and the characters were only in isolated forms.

The third row represents carried out experiemnts on the merged dataset Hijja-AHCD. Compared with with the results on Hijja dataset alone in the first row, it can be seen that merging Hijja with AHCD improved the predection accuracy in our model, Ahmed, et al. [24], and EfficientNetV0.

These results show the challenges evolving from the age difference between participants in each dataset. In Hijja, they were all children, and children's handwritings could be very challenging to understand even for the human eye, so to capture more features and help the model learn more possible ways of writing each character, merging Hijja with AHCD experimentally helped the model learn better. We assumed that showing the model more variations of handwritings is more effective and realistic for our problem than enlarging the size of Hijja dataset with usual augmentation techniques like zooming or rotating existing images, hence we decided to merge with another similar dataset with a different age segmant instead of generating new images from the same dataset.

Lastly, we introduced a new approach which is based on a prior filteration step using character strokes. We calculated the averaged predection accuracy on all 4 character groups as shown previously in **Table 2**, and is reported in the last column in **Table 3** as **multi-model**. The averaged predection accuracy outperformed all the single model experiments and transfer learning on Hijja-AHCD dataset. The classification reports for each of the models in each character group are shown in **Tables A 4 to 7** in the appendix. In order to make use of this method in a real-life application, there should be a layer to filter the number of strokes in an image prior to prediction, and based on that number, which model is used for prediction will be decided. Such thing is doable in online writing applications where the strokes can be counted as the number of times the user interacts with the screen using their fingers or a stylus. Experimentally, we split the test set as we did with the training set for evaluation.

During the evaluation, we noticed some factors of which can be said as challenges facing the CNN models with handwritten Arabic characters. First, as can be seen in **Figure 2**, the model misclassified the character Qaaf/ق as Faa'/ف, we provide the image of both characters clearly shown in Figure 9 for comparison. We can see that Faa' has 1 dot, and Qaaf has 2 dots, while the shape of the character itself is more curved than Faa', but many handwriting styles could make the main part of both look very similar, leaving the only difference in the dots above the curve. The image we showed the model has the 2 dots very unclear due to scanning and data preprocessing, this can be seen as a problem from two sides. Firstly, with the traditional data collecting method by scanning written characters from paper, and due to cleaning and preprocessing, it loses some quality.

Secondly, the Arabic characters are a challenging problem since some characters look very similar, and even more challenging with the diverse handwriting styles.

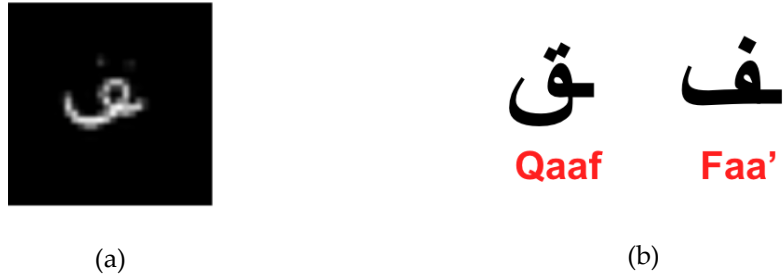

(a)                                              (b)

*Figure 6: (a) input image of Qaaf on its final shape misclassified as Faa' (b) printed Qaaf and Faa'.*

Furthermore, the model made similar misclassifications on characters shown in **Figure 7**. All of them were part of the Hijja dataset, and they are written in the medial shape, that is, they are written as if they were in the middle of a word. The model misclassified them with similar characters written in the exact same way, also in the middle position. It's challenging even for us to classify these characters, and with the model reaching a considerable good prediction accuracy over most cases with isolated characters, it's still unable to overcome some challenges with other positions.

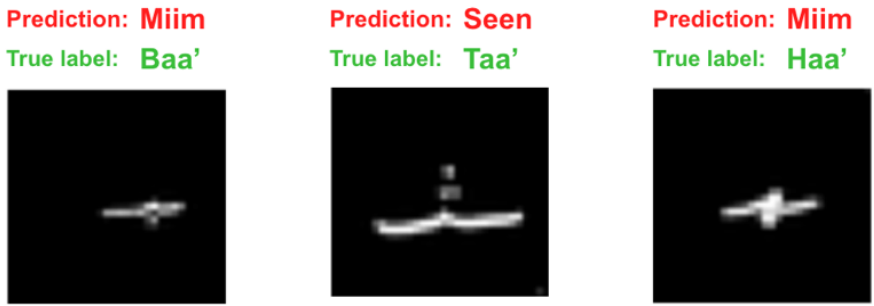

*Figure 7: more misclassifications made by the model on Hijja dataset due to similarity between isolated and connected characters*

Our proposal of the multi-model stroke-based solution opens the door for a new way of solving classification problems in the Arabic handwriting with the idea of a prior filtration process to overcome the shortages mentioned earlier.

## 5 Conclusion and Future Work

The Arabic language introduces more challenges in the fields of Deep Learning, more studies are being carried out yet not enough to reach the advanced results in Latin languages. In this research, we discuss some approaches to get higher accuracy in Arabic handwritten character recognition using Deep Learning over 2 recent datasets, one of which was written by children and introduces even higher complex patterns, the Hijja dataset. We used augmentation with another dataset, AHCD, and strokes approach on the same model. Our model reached higher accuracy on both Hijja and AHCD than the baseline, moreover, augmenting Hijja with AHCD helped raise the accuracy further than on Hijja alone, which tells that the dataset made learning the handwritten characters patterns harder for the model.

We compared our model with transfer learning using EfficientNetV0 and one of the recent proposed models that gained high results on Arabaic handwritten charcters in the litrature, nevertheless, our model outperform both their results and transfer learning. Furthermore, we also compared the multi-model approach with both the results from single models in (A) and transfer learning (C) on Hijja-AHCD and it outperformed both.

As future work, we plan to test our model in a real application as a proof of concept which could further reveal if these experiments are applicable under real conditions outside the testing environment. Particularly, using our model in an online writing application, where we present the model with a different type of input, and we anticipate it to perform well since images extracted from online handwriting are more clear than offline images, which could help the model recognize the features easily. In addition, we see that creating a new dataset based on online writing could further help improve the accuracy of our model in such applications.

As we focused more on the complex patterns of children's handwriting, the applications of this model in educational apps to teach dictation for children are a good use case for this research.

## Appendix A

This appendix contains the full classification reports of all experiment in **Table 3**.

**Table A 1.** full classification report of our model on Hijja dataset

| Label | precision | recall | f1-score |
|---|---|---|---|
| **Alif** ا | 0.99 | 0.99 | 0.99 |
| **Baa'** ب | 0.94 | 0.96 | 0.95 |
| **Taa'** ت | 89 | 0.93 | 0.91 |
| **Thaa'** ث | 0.95 | 0.92 | 0.93 |
| **Jiim** ج | 0.94 | 0.96 | 0.95 |
| **Haa'** ح | 0.9 | 0.81 | 0.85 |
| **Khaa'** خ | 0.9 | 0.88 | 0.89 |
| **Daal** د | 0.76 | 0.83 | 0.79 |
| **Dhaal** ذ | 0.77 | 0.71 | 0.74 |
| **Raa'** ر | 0.93 | 0.91 | 0.92 |
| **Zaay** ز | 0.92 | 0.88 | 0.9 |
| **Seen** س | 0.95 | 0.96 | 0.96 |
| **Sheen** ش | 0.98 | 0.95 | 0.97 |
| **Saad** ص | 0.86 | 0.94 | 0.9 |
| **Daad** ض | 0.92 | 0.88 | 0.9 |
| **Taa'** ط | 0.95 | 0.91 | 0.93 |
| **Zaa'** ظ | 0.93 | 0.96 | 0.94 |
| **Aayn** ع | 0.82 | 0.87 | 0.84 |
| **Ghayn** غ | 0.92 | 0.84 | 0.88 |
| **Faa'** ف | 0.82 | 0.87 | 0.84 |
| **Qaaf** ق | 0.92 | 0.93 | 0.93 |
| **Kaff** ك | 0.91 | 0.89 | 0.9 |
| **Lamm** ل | 0.92 | 0.91 | 0.92 |
| **Miim** م | 0.91 | 0.93 | 0.92 |
| **Nuun** ن | 0.79 | 0.87 | 0.83 |
| **Haa'** ه | 0.92 | 0.88 | 0.9 |
| **Waaw** و | 0.92 | 0.94 | 0.93 |
| **Yaa'** ي | 0.97 | 0.93 | 0.95 |
| **Hamza** ء | 0.84 | 0.87 | 0.86 |
| accuracy | | | 0.91 |
| macro avg | 0.90 | 0.90 | 0.90 |
| weighted avg | 0.91 | 0.91 | 0.91 |

**Table A 2.** full classification report of our model on AHCD dataset

| Label | precision | recall | f1-score |
|---|---|---|---|
| **Alif** ا | 0.99 | 1 | 1 |
| **Baa'** ب | 0.99 | 0.99 | 0.99 |
| **Taa'** ت | 0.93 | 0.96 | 0.94 |
| **Thaa'** ث | 0.96 | 0.95 | 0.95 |
| **Jiim** ج | 0.99 | 0.97 | 0.98 |
| **Haa'** ح | 0.97 | 0.97 | 0.97 |
| **Khaa'** خ | 1 | 0.98 | 0.99 |
| **Daal** د | 0.93 | 0.98 | 0.96 |
| **Dhaal** ذ | 0.94 | 0.95 | 0.95 |
| **Raa'** ر | 0.96 | 0.97 | 0.96 |
| **Zaay** ز | 0.99 | 0.92 | 0.95 |
| **Seen** س | 1 | 0.99 | 1 |
| **Sheen** ش | 0.98 | 1 | 0.99 |
| **Saad** ص | 0.96 | 1 | 0.98 |
| **Daad** ض | 0.99 | 0.96 | 0.97 |
| **Taa'** ط | 0.96 | 0.98 | 0.97 |
| **Zaa'** ظ | 0.99 | 0.96 | 0.97 |
| **Aayn** ع | 0.98 | 0.99 | 0.98 |
| **Ghayn** غ | 1 | 0.98 | 0.99 |
| **Faa'** ف | 0.97 | 0.96 | 0.96 |
| **Qaaf** ق | 0.95 | 0.95 | 0.95 |
| **Kaff** ك | 0.97 | 0.97 | 0.97 |
| **Lamm** ل | 0.98 | 1 | 0.99 |
| **Miim** م | 0.99 | 0.98 | 0.99 |
| **Nuun** ن | 0.97 | 0.94 | 0.96 |
| **Haa'** ه | 0.98 | 0.98 | 0.98 |
| **Waaw** و | 0.94 | 0.98 | 0.96 |
| **Yaa'** ي | 1 | 0.97 | 0.99 |
| **Hamza** ء | | | 0.97 |
| accuracy | 0.97 | 0.97 | 0.97 |
| macro avg | 0.97 | 0.97 | 0.97 |

**Table A 3.** full classification report of our model on Hijja-AHCD dataset

| Label | precision | recall | f1-score |
|---|---|---|---|
| **Alif** ا | 0.99 | 0.99 | 0.99 |
| **Baa'** ب | 0.97 | 0.97 | 0.97 |
| **Taa'** ت | 0.93 | 0.93 | 0.93 |

| Label | | | |
|---|---|---|---|
| **Thaa'** ث | 0.95 | 0.93 | 0.94 |
| **Jiim** ج | 0.96 | 0.95 | 0.96 |
| **Haa'** ح | 0.89 | 0.93 | 0.91 |
| **Khaa'** خ | 0.95 | 0.93 | 0.94 |
| **Daal** د | 0.84 | 0.87 | 0.86 |
| **Dhaal** ذ | 0.81 | 0.85 | 0.83 |
| **Raa'** ر | 0.94 | 0.92 | 0.93 |
| **Zaay** ز | 0.93 | 0.89 | 0.91 |
| **Seen** س | 0.97 | 0.97 | 0.97 |
| **Sheen** ش | 0.98 | 0.96 | 0.97 |
| **Saad** ص | 0.9 | 0.94 | 0.92 |
| **Daad** ض | 0.92 | 0.92 | 0.92 |
| **Taa'** ط | 0.93 | 0.94 | 0.94 |
| **Zaa'** ظ | 0.97 | 0.92 | 0.94 |
| **Aayn** ع | 0.89 | 0.88 | 0.88 |
| **Ghayn** غ | 0.94 | 0.89 | 0.91 |
| **Faa'** ف | 0.82 | 0.92 | 0.87 |
| **Qaaf** ق | 0.94 | 0.9 | 0.92 |
| **Kaff** ك | 0.91 | 0.93 | 0.92 |
| **Lamm** ل | 0.95 | 0.92 | 0.93 |
| **Miim** م | 0.93 | 0.95 | 0.94 |
| **Nuun** ن | 0.86 | 0.87 | 0.86 |
| **Haa'** ه | 0.93 | 0.9 | 0.92 |
| **Waaw** و | 0.94 | 0.97 | 0.96 |
| **Yaa'** ي | 0.97 | 0.96 | 0.97 |
| **Hamza** ء | 0.87 | 0.86 | 0.86 |
| **accuracy** | | | 0.93 |
| **macro avg** | 0.92 | 0.92 | 0.92 |
| **weighted avg** | 0.93 | 0.93 | 0.93 |

**Table A 4.** full classification report of multi model experiment, group 1

| Label | precision | recall | f1-score |
|---|---|---|---|
| **Alif** ا | 0.99 | 0.99 | 0.99 |
| **Haa'** ح | 0.89 | 0.93 | 0.91 |
| **Daal** د | 0.89 | 0.82 | 0.85 |
| **Raa'** ر | 0.94 | 0.93 | 0.94 |
| **Seen** س | 0.97 | 0.95 | 0.96 |
| **Saad** ص | 0.93 | 0.93 | 0.93 |
| **Taa'** ط | 0.98 | 0.94 | 0.96 |
| **Aayn** ع | 0.89 | 0.91 | 0.9 |
| **Lamm** ل | 0.95 | 0.94 | 0.94 |
| **Miim** م | 0.92 | 0.94 | 0.93 |

| Label | | precision | recall | f1-score |
|---|---|---|---|---|
| Haa' | هـ | 0.92 | 0.93 | 0.92 |
| Waaw | و | 0.93 | 0.97 | 0.95 |
| Hamza | ء | 0.92 | 0.94 | 0.93 |
| accuracy | | | | 0.94 |
| macro avg | | 0.93 | 0.93 | 0.93 |
| weighted avg | | 0.94 | 0.94 | 0.94 |

**Table A 5.** full classification report of multi model experiment, group 2

| Label | | precision | recall | f1-score |
|---|---|---|---|---|
| Baa' | ب | 0.97 | 0.97 | 0.97 |
| Taa' | ت | 0.92 | 0.93 | 0.92 |
| Thaa' | ث | 0.93 | 0.95 | 0.94 |
| Jiim | ج | 0.98 | 0.98 | 0.98 |
| Khaa' | خ | 0.93 | 0.94 | 0.93 |
| Dhaal | ذ | 0.83 | 0.84 | 0.84 |
| Zaay | ز | 0.95 | 0.94 | 0.95 |
| Sheen | ش | 0.98 | 0.96 | 0.97 |
| Daad | ض | 0.95 | 0.94 | 0.94 |
| Zaa' | ظ | 0.96 | 0.98 | 0.97 |
| Ghayn | غ | 0.92 | 0.89 | 0.91 |
| Faa' | ف | 0.86 | 0.88 | 0.87 |
| Qaaf | ق | 0.93 | 0.92 | 0.93 |
| Kaff | ك | 0.95 | 0.95 | 0.95 |
| Nuun | ن | 0.87 | 0.88 | 0.88 |
| Yaa' | ي | 0.97 | 0.97 | 0.97 |
| accuracy | | | | 0.93 |
| macro avg | | 0.93 | 0.93 | 0.93 |
| weighted avg | | 0.93 | 0.93 | 0.93 |

**Table A 6.** full classification report of multi model experiment, group 3

| Label | | precision | recall | f1-score |
|---|---|---|---|---|
| Taa' | ت | 0.99 | 0.98 | 0.98 |
| Zaa' | ظ | 0.99 | 0.98 | 0.99 |
| Qaaf | ق | 0.98 | 0.98 | 0.98 |
| Yaa' | ي | 0.99 | 1.00 | 0.99 |
| accuracy | | | | 0.99 |
| macro avg | | 0.99 | 0.99 | 0.99 |
| weighted avg | | 0.99 | 0.99 | 0.99 |

**Table A 7.** full classification report of multi model experiment, group 4

| Label | precision | recall | f1-score |
|---|---|---|---|
| **Thaa' ثـ** | 0.98 | 0.99 | 0.99 |
| **Sheen شـ** | 0.99 | 0.98 | 0.99 |
| **accuracy** | | | 0.99 |
| **macro avg** | 0.99 | 0.99 | 0.99 |
| **weighted avg** | 0.99 | 0.99 | 0.99 |